\documentclass[journal]{IEEEtran}
\ifCLASSINFOpdf
\else
\fi
\usepackage{amssymb}
\usepackage{hyperref}
\usepackage{amsmath,bm}
\usepackage{algorithm}
\usepackage{algpseudocode}
\usepackage{cite}
\usepackage{subfig}
\usepackage{tabularx}
\usepackage{tabulary}
\usepackage{mathrsfs}
\usepackage{graphicx}
\usepackage{epstopdf}
\usepackage{multirow}
\usepackage{makecell}
\usepackage{booktabs}

\begin{document}
\title{Correntropy-Based Logistic Regression \\ with Automatic Relevance Determination for \\ Robust Sparse Brain Activity Decoding}
\author{Yuanhao~Li,
	Badong~Chen,~\IEEEmembership{Senior~Member,~IEEE,}
	\\
	Yuxi~Shi,
	Natsue~Yoshimura,
	and~Yasuharu~Koike
	\thanks{This work was supported in part by the Japan Society for the Promotion of Science (JSPS) KAKENHI under Grant 19H05728, in part by the Japan Science and Technology Agency (JST) Support for the Pioneering Research Initiated by Next Generation (SPRING) under Grant JPMJSP2106, and in part by the National Natural Science Foundation of China under Grant U21A20485 and Grant 61976175. \emph{(Corresponding author: Yuanhao Li.)}}
	\thanks{Yuanhao Li, Yuxi Shi, Natsue Yoshimura, and Yasuharu Koike are with the Institute of Innovative Research, Tokyo Institute of Technology, Yokohama 226-8503, Japan. (correspondence e-mail: li.y.ay@m.titech.ac.jp)}
	\thanks{Badong Chen is with the Institute of Artificial Intelligence and Robotics, Xi'an Jiaotong University, Xi'an 710049, China.}} 
\maketitle

\begin{abstract}
Recent studies have utilized sparse classifications to predict categorical variables from high-dimensional brain activity signals to expose human's intentions and mental states, selecting the relevant features automatically in the model training process. However, existing sparse classification models will likely be prone to the performance degradation which is caused by noise inherent in the brain recordings. To address this issue, we aim to propose a new robust and sparse classification algorithm in this study. To this end, we introduce the correntropy learning framework into the automatic relevance determination based sparse classification model, proposing a new correntropy-based robust sparse logistic regression algorithm. To demonstrate the superior brain activity decoding performance of the proposed algorithm, we evaluate it on a synthetic dataset, an electroencephalogram (EEG) dataset, and a functional magnetic resonance imaging (fMRI) dataset. The extensive experimental results confirm that not only the proposed method can achieve higher classification accuracy in a noisy and high-dimensional classification task, but also it would select those more informative features for the decoding scenarios. Integrating the correntropy learning approach with the automatic relevance determination technique will significantly improve the robustness with respect to the noise, leading to more adequate robust sparse brain decoding algorithm. It provides a more powerful approach in the real-world brain activity decoding and the brain-computer interfaces.      
\end{abstract}


\section{Introduction}
\label{sec:introduction}
\IEEEPARstart{B}{rain} activity decoding technology has been increasingly driving the developments of the brain-computer interface (BCI) systems, which aims to reveal the human's mental states \cite{wolpaw2002brain,haynes2006decoding}. Non-invasive brain activity recording approaches, such as multi-channel electroencephalogram (EEG) and functional magnetic resonance imaging (fMRI), are prevalently employed to develop advanced brain decoding paradigms, mainly due to their ease of operation and surgery-free merit. Numerous brain decoding scenarios have been investigated in the literature. For example, EEG-based motor imagery is a canonical task, which aims to forecast the motion patterns imagined by the subjects \cite{padfield2019eeg}. The P300 event-related potential, which could be detected from EEG signals, has proved to be a promising way to control a speller BCI system \cite{krusienski2006comparison}. Steady-state visual evoked potentials were used to predict visual stimulus through frequency domain analysis \cite{jia2010frequency}. As for fMRI-based studies, the visual perceptions of different edge orientations were successfully decoded in \cite{kamitani2005decoding}. Further, a more challenging issue, constraint-free visual image reconstruction, was solved in \cite{miyawaki2008visual}. Moreover, memory detection has proved to be practicable with fMRI signals \cite{peth2015memory}.

To expose the information and patterns inherent in the brain recording signals, in particular to predict categorical variables, such as whether the subject is imaging left or right, one usually exploits multivariate classification models \cite{haynes2006decoding,norman2006beyond,pereira2009machine}, where the logistic regression has been widely utilized in the literature \cite{menard2002applied}. Although the logistic regression model performs well for a generic classification task, it is vulnerable to the performance degradation caused by over-fitting, especially when the number of features is larger than that of samples, i.e. high-dimensional problem. Brain decoding tasks are considerably prone to high- dimensional problem which may result from multiple channels and wide bandwidth of EEG signals, and thousands of voxels in fMRI signals \cite{muller2004machine}.

To address the over-fitting problem in brain decoding, many studies have investigated various strategies. A primary method to avoid over-fitting is to utilize $L_2$-norm regularization so that the model parameters are shrunk towards zero and the variance of the classifier is decreased as a result \cite{salehi2019impact}. Another approach is to select a subset of the features so that the dimension would be reduced, which is commonly achieved by $L_1$-regularization to realize a sparse model \cite{tibshirani1996regression,ng2004feature}. The sparse model has been increasingly attractive for brain decoding because the selected features could indicate the spatio-temporal patterns relevant to the specific cognitive tasks \cite{yamashita2008sparse,ryali2010sparse,satake2018sparse,ganesh2008sparse}.

An alternative method to achieve a sparse classifier is to use a sparse prior distribution on the model parameters and update the classification model from Bayesian perspective. In fact, the $L_1$-regularization is equivalent to employing a Laplacian prior distribution \cite{tibshirani1996regression}. Compared to regularization, an advantage of using a sparse prior distribution is that, one does not need to adjust the regularization parameter manually \cite{figueiredo2003adaptive}. Notably, the automatic relevance determination (ARD) technique \cite{mackay1992bayesian} was introduced into the logistic regression classifier \cite{yamashita2008sparse}, which is a hierarchical sparse prior and has proved to be more adequate than the Laplacian prior (equally $L_1$-regularization) for feature selection \cite{wipf2007new}. The ARD-based sparse logistic regression \cite{yamashita2008sparse} (SLR) has been widely employed for brain decoding, including EEG decoding \cite{lisi2014decoding,lisi2015eeg,ganesh2018utilizing,shi2021galvanic}, fMRI decoding \cite{miyawaki2008visual,shibata2011perceptual,yahata2016small,horikawa2017generic}, and current source density analysis \cite{morioka2014decoding,yoshimura2016decoding,mejia2018decoding}.

However, the decoding performance of SLR may be limited, since it assumes the Bernoulli distribution on the data in binary classification, of which the validity can be degenerated hugely in a noisy classification task \cite{li2021restricted}. It is noted that non-invasive brain recordings are known to be prone to noises. For example, blinks and eye movements deteriorate EEG signals \cite{ball2009signal}, while fMRI suffers both physiological noises and motions \cite{liu2016noise}. Even though many preprocessing methods were developed to isolate noises, such as the blind source separation by the independent component analysis \cite{jung2000removing}, one can hardly ensure that all noises are well removed by preprocessing.

In the present study, we aim to investigate a robust version of the SLR algorithm, which can suppress the negative effects caused by noises inherent in the brain measurement signals. To this end, we introduce the correntropy learning framework into the ARD-based sparse model. Correntropy learning, developed in information theoretic learning (ITL) \cite{principe2010information}, exhibits inspiring robustness for various machine learning and signal processing tasks, including regression \cite{liu2007correntropy,feng2015learning}, classification \cite{singh2014c,xu2016robust}, principal component analysis \cite{he2011robust}, and so on.

The remainder of this paper is organized as follows. Section \ref{sec:related_works} briefly reviews the generic logistic regression algorithm, the hierarchical ARD prior distribution, and the ARD-based SLR algorithm. In Section \ref{sec:method}, we introduce the correntropy learning approach, and then propose a new robust sparse classification algorithm by integrating the correntropy learning and the ARD prior distribution. In Section \ref{sec:exp}, to demonstrate the robustness and feature selection capability of the proposed algorithm, we present extensive experimental results with a synthetic dataset, an EEG dataset, and an fMRI dataset, respectively. In addition, we provide some discussions in Section \ref{sec:disc}. Finally, the paper is concluded in Section \ref{sec:con}.

\section{Preliminaries}
\label{sec:related_works}

Before presenting the correntropy-based robust sparse classification for brain activity decoding, we review in this section the logistic regression classification model, the ARD hierarchical sparse prior, and the ARD-based SLR algorithm.

\subsection{Logistic Regression}
\label{subsec:lr}
Logistic regression is a widely used probability-based classification model. A linear discriminant function that recognizes two different classes can be expressed by a weighted summation of each input feature
\begin{equation}
\label{equ:dis_function}
f(\mathbf{x},\omega)=\sum_{d=1}^D{\omega_dx_d}+\omega_0
\end{equation}
where $\mathbf{x}=(x_1,x_2,\cdots,x_D)\in \mathbb{R}^D$ denotes a $D$-dimensional input sample to be classified, while $\omega=(\omega_0,\omega_1,\omega_2,\cdots,\omega_D)$ represents the logistic regression model parameter, with a bias term contained. In the $\{$0,1$\}$-label context, logistic regression computes the probability that an input sample belongs to class 1 through the \emph{sigmoid} function as
\begin{equation}
\label{equ:sig_function}
y\triangleq P(t=1|\mathbf{x},\omega)=\frac{1}{1+\exp(-f(\mathbf{x},\omega))}
\end{equation}
where $t$ is the label. Assuming the Bernoulli distribution leads to the opposite probability for class 0 as $P(t=0|\mathbf{x},\omega)=1-y$. Given a finite dataset $\{(\mathbf{x}_n,t_n) \}_{n=1}^{N}$ the likelihood function is
\begin{equation}
\begin{split}
\label{equ:likelihood_function}
&P(t_1,t_2,\cdots,t_N|\mathbf{x}_1,\mathbf{x}_2,\cdots,\mathbf{x}_N,\omega)\; \\
=&\prod_{n=1}^{N}{P(t_n|\mathbf{x}_n,\omega)}=\prod_{n=1}^{N}{y_n^{t_n}(1-y_n)^{1-t_n}}\; \\
\end{split}
\end{equation}
in which $y_n$ represents the probability of the \emph{n}-th sample for class 1 by
\begin{equation}
\label{equ:p_n}
y_n\triangleq P(t_n=1|\mathbf{x}_n,\omega)=\frac{1}{1+\exp(-f(\mathbf{x}_n,\omega))}
\end{equation}
Maximizing the likelihood probability in (\ref{equ:likelihood_function}) with an observed dataset is equal to maximizing the logarithmic form, yielding
\begin{equation}
\label{equ:lr}
\omega^*=\text{argmax}\sum_{n=1}^{N}{(t_n\log y_n+(1-t_n)\log (1-y_n))}
\end{equation}
in which $\omega^*$ indicates the optimal model parameter. After one trains the logistic regression classifier, for a new testing sample $\mathbf{x}$, the label is predicted as class 1 if $f(\mathbf{x},\omega^*)>0$ (or equally $P(t=1|\mathbf{x},\omega^*)>0.5$), otherwise as class 0.

\subsection{Automatic Relevance Determination}
\label{subsec:ard}
The maximum likelihood estimation (MLE), as is mentioned above, only takes into account the distribution for the data. On the other hand, one could assume a prior distribution for model parameter moreover.

ARD is a sophisticated sparse prior with a hierarchical form, proved to exhibit superior feature selection capability than the Laplacian prior \cite{wipf2007new}. ARD employs the anisotropic Gaussian distribution for each model parameter, that each parameter has an individual variance
\begin{equation}
\label{equ:ard1}
P(\omega_d|\alpha_d) = \mathcal{N}(0,\alpha_d^{-1}) \qquad d=0,1,\cdots,D
\end{equation}
in which $\mathcal{N}$ is the univariate Gaussian distribution, and $\alpha_d$ is called the relevance parameter, controlling the possible range of the corresponding model parameter. Each hyper-parameter is further assumed with the non-informative prior distribution
\begin{equation}
\label{equ:ard2}
P_0(\alpha_d) = \alpha_d^{-1} \qquad d=0,1,\cdots,D
\end{equation}
in a full-Bayesian formulation.

\subsection{ARD-Based Sparse Logistic Regression}
\label{subsec:slr}
The ARD-based SLR algorithm estimates the model parameters and the relevance parameters by a maximum a posteriori (MAP) method. The posterior distribution for $\omega$ is calculated
\begin{equation}
\label{equ:post_distribution}
P(\omega|{T},{X})= \frac{\int P({T}|\omega,{X})P(\omega|\alpha)P_0(\alpha)\text{d}\alpha}{\int P({T}|\omega,{X})P(\omega|\alpha)P_0(\alpha)\text{d}\alpha\text{d}\omega}
\end{equation}
where ${T}$ and ${X}$ denote the labels and attributes of the observed dataset, respectively, and $\alpha=(\alpha_1,\alpha_2,\cdots,\alpha_D)$ is a collection of the relevance parameters. The integration in (\ref{equ:post_distribution}) is, however, difficult to calculate analytically. Hence, variational inference \cite{blei2017variational} was introduced for approximation, defining the following free energy on $Q(\omega,\alpha)$ to approximate the true joint posterior distribution $P(\omega,\alpha|{T},{X})$ by
\begin{equation}
\label{equ:free_energy_1}
F(Q(\omega,\alpha))= \int Q(\omega,\alpha)\log\frac{P({T},\omega,\alpha)}{Q(\omega,\alpha)}\text{d}\alpha \text{d}\omega 
\end{equation}
where the dependency on ${X}$ is omitted for simplicity. The free energy $F(Q(\omega,\alpha))$ would be maximized when and only when $Q(\omega,\alpha)$ is equal to $P(\omega,\alpha|{T},{X})$. To maximize $F(Q(\omega,\alpha))$, one further assumes the conditional independence between $\omega$ and $\alpha$ as $Q(\omega,\alpha)=Q(\omega)Q(\alpha)$, leading to
\begin{equation}
\label{equ:free_energy_2}
F(Q(\omega)Q(\alpha))= \int Q(\omega)Q(\alpha)\log\frac{P({T},\omega,\alpha)}{Q(\omega)Q(\alpha)}\text{d}\alpha \text{d}\omega 
\end{equation}
Thus one can maximize the free energy alternately with respect to $Q(\omega)$ and $Q(\alpha)$, respectively, with the following updates 
\begin{equation}
\begin{split}
\label{equ:slr}
&\omega\text{-step:} \quad \log Q(\omega)=\left< \log P({T},\omega,\alpha)\right>_{Q(\alpha)}+const \; \\
&\alpha\text{-step:} \quad \log Q(\alpha)=\left< \log P({T},\omega,\alpha)\right>_{Q(\omega)}+const \; \\
\end{split}
\end{equation}
in which $\left< x\right>_{Q(x)}$ denotes the expectation of $x$ with respect to the probability distribution $Q(x)$.

Despite the exceptional capability for feature selection, SLR suffers the significant performance degradation resulting from noises in practice, because its foundation is the assumption of Bernoulli distribution for the likelihood, which was proved to exhibit poor robustness with respect to noises \cite{li2021restricted}. Hence, the optimization process for $\omega$-step will be degenerated in a noisy classification task. The deteriorated solution for $\omega$-step would further destabilize the calculation for the relevance parameters in $\alpha$-step, which provides imprecise $\alpha_d$ for the next $\omega$-step in turn. Thus, both model parameter $\omega$ and relevance parameter $\alpha$ are deviated from the desired values.

\section{Method}
\label{sec:method}

\subsection{Correntropy Learning}
\label{subsec:mcc}
Correntropy, developed as a generalized correlation function for stochastic process, has been further generalized to measure the similarity between two arbitrary variables in a reproducing kernel Hilbert space \cite{liu2007correntropy}. Correntropy has a close relation with the information potential of the quadratic Renyi's entropy \cite{principe2010information}, in which the Parzen's window method is employed to estimate the probability density of the data distribution. Concerning two arbitrary random variables $A$ and $B$, the correntropy is defined
\begin{equation}
\label{equ:correntropy}
\mathcal{V}(A,B)=E[k(A-B)]
\end{equation}
in which $E$ is the mathematical expectation while $k$ is a kernel function. One usually uses the Gaussian kernel function which is denoted by $k_\sigma(x)\triangleq \exp(-x^2/2\sigma^2)$ with kernel bandwidth $\sigma$. Thus, $N$ observations of the variables $A$ and $B$, in practice, bring about the following empirical estimation of correntropy
\begin{equation}
\begin{split}
\label{equ:correntropy_est}
\hat{\mathcal{V}}(A,B)&=\frac{1}{N}\sum_{n=1}^{N}{k_\sigma(a_n-b_n)} \; \\
&=\frac{1}{N}\sum_{n=1}^{N}{\exp(-\frac{(a_n-b_n)^2}{2\sigma^2})} \; \\
\end{split}
\end{equation}
If one desires to maximize the similarity between two variables in a machine learning task, such as between the predicted and the target output, one can train a learning machine to maximize their correntropy, which is called as the maximum correntropy criterion (MCC) \cite{liu2007correntropy}. Employing a correntropy-based learning criterion exhibits many benefits. First, the value of correntropy is mainly determined by the Gaussian kernel function $k_\sigma$ along $A=B$, which means it is a local similarity measure and could alleviate the negative effect of large deviation caused by noise. It was also proved to extract more information from the dataset because it involves all the even-numbered statistical moments. In addition, it has a close relation with the $m$-estimation. One could find a comprehensive explanation of correntropy in \cite{liu2007correntropy}.

\subsection{Correntropy-Based Robust Sparse Classification}
\label{subsec:cslr}
The MAP calculation in (\ref{equ:post_distribution}), when one omits the dependency on ${X}$, could be rewritten by integration as
\begin{equation}
\label{equ:max_post_distribution_1}
P(\omega|{T})=\frac{P({T}|\omega)P_0(\omega)}{P({T})}\propto P({T}|\omega)P_0(\omega)
\end{equation}
where $P({T})$ is a constant that is usually called evidence, and $P_0(\omega)=\int P(\omega|\alpha)P_0(\alpha)\text{d}\alpha$ indicates the prior distribution for $\omega$ by integrating out $\alpha$. Since the logarithm is a monotonically increasing function, MAP is equal to
\begin{equation}
\begin{split}
\label{equ:max_post_distribution_2}
\max\ & P(\omega|{T})  \; \\
\Leftrightarrow \max\ & \log P(\omega|{T}) \; \\
\Leftrightarrow \max\ & \log P({T}|\omega)+\log P_0(\omega) \; \\
\end{split}
\end{equation}
The poor robustness of the original SLR algorithm results from the log likelihood item $\log P({T}|\omega)$. Since likelihood measures the probability that the prediction is equal to the target output, comparably, we could employ the local correntropy instead to evaluate the similarity between prediction and desired output, motivated by the admirable robustness of correntropy learning criterion. Therefore, we use the correntropy similarity between the prediction and target $\mathcal{V}({T},{Y})$, in which $Y$ is the collection of the predicted probability $y$, to substitute the non-robust log likelihood $\log P({T}|\omega)$. The superior robustness of correntropy $\mathcal{V}({T},{Y})$, compared to log likelihood $\log P({T}|\omega)$, was verified on ordinary logistic regression model and radial basis function network with extensive experimental results \cite{li2021restricted,singh2014c}.

To coordinate the correntropy item $\mathcal{V}({T},{Y})$ to the Bayesian derivation of SLR described in Section \ref{subsec:slr}, we could rewrite the free energy maximization in (\ref{equ:free_energy_2}) as
\begin{equation}
\begin{split}
\label{equ:free_energy_3}
\max &\int Q(\omega)Q(\alpha)\times \; \\
&(\log P({T},\omega,\alpha)-\log Q(\omega)-\log Q(\alpha))\text{d}\alpha \text{d}\omega \; \\
\end{split}
\end{equation}
where the log joint distribution $\log P({T},\omega,\alpha)$ is decomposed
\begin{equation}
\label{equ:log_joint_distribution}
\log P({T},\omega,\alpha)=\log P({T}|\omega)+\log P(\omega|\alpha)+\log P_0(\alpha)
\end{equation}
from which one can also find the non-robust $\log P({T}|\omega)$. Here we propose the correntropy-based pseudo log joint distribution $\log P_c({T},\omega,\alpha)$, which is defined by
\begin{equation}
\label{equ:log_joint_distribution_c}
\log P_c({T},\omega,\alpha)\triangleq\mathcal{V}({T},{Y})+\log P(\omega|\alpha)+\log P_0(\alpha)
\end{equation}
where $P_c({T},\omega,\alpha)$ is called pseudo joint distribution since one finds its integration over all values is not equal to one.

By defining the novel pseudo joint distribution $P_c({T},\omega,\alpha)$, we reformulate the free energy maximization as 
\begin{equation}
\begin{split}
\label{equ:free_energy_new}
\max &\int Q(\omega)Q(\alpha)\times \; \\
&(\log P_c({T},\omega,\alpha)-\log Q(\omega)-\log Q(\alpha))\text{d}\alpha \text{d}\omega \; \\
\end{split}
\end{equation}
Similarly, one can acquire the following alternate optimization 
\begin{equation}
\begin{split}
\label{equ:cslr}
&\omega\text{-step:} \quad \log Q_c(\omega)=\left< \log P_c({T},\omega,\alpha)\right>_{Q_c(\alpha)}+const \; \\
&\alpha\text{-step:} \quad \log Q_c(\alpha)=\left< \log P_c({T},\omega,\alpha)\right>_{Q_c(\omega)}+const \; \\
\end{split}
\end{equation}
Computing the expectation and omitting the constant in $\omega$-step, one obtains
\begin{equation}
\label{equ:cslr_w_a}
\log Q_c(\omega)= \frac{1}{N}\sum_{n=1}^{N}{\exp(-\frac{(t_n-y_n)^2}{2\sigma^2})}-\frac{1}{2}\omega^t\varLambda\omega
\end{equation}
where $\varLambda=\text{diag}(\alpha_1,\cdots,\alpha_D)$. One finds that, however, $\omega$-step is analytically intractable since the likelihood and the prior are not conjugate. We further employ the Laplacian approximation to $\log Q_c(\omega)$, a quadratic approximation at the MAP estimate, denoted by $\omega^*$, yielding
\begin{equation}
\label{equ:cslr_w_gau}
\log Q_c(\omega)\approx \log Q_c(\omega^*)-\frac{1}{2}(\omega-\omega^*)^tH_c(\omega^*)(\omega-\omega^*)
\end{equation}
where $H_c(\omega^*)$ is the negative Hessian matrix of $\log Q_c(\omega)$ at $\omega^*$. Thus, $Q_c(\omega)$ is approximated with a Gaussian distribution $\mathcal{N}(\omega^*,S_c)$, where $S_c\triangleq H_c(\omega^*)^{-1}$. The gradient of $\log Q_c(\omega)$ with respect to model parameter is given by
\begin{equation}
	\begin{split}
		\label{equ:cslr_g}
		&\frac{\partial\log Q_c(\omega)}{\partial\omega} \; \\
		&=\frac{1}{N\sigma^2}\sum_{n=1}^{N}{\exp(-\frac{e_n^2}{2\sigma^2})e_ny_n(1-y_n)\mathbf{x}_n}-\varLambda\omega \; \\
	\end{split}
\end{equation}
where $e_n=t_n-y_n$ is the prediction error. The Hessian matrix of $\log Q_c(\omega)$ is given in (\ref{equ:cslr_h}).

In $\alpha$-step, given $Q_c(\omega)=\mathcal{N}(\omega^*,S_c)$, one can obtain
\begin{equation}
\label{equ:slr_a}
\log Q_c(\alpha)= -\frac{1}{2}\sum_{d=1}^{D}{(\alpha_d(\omega_d^{*2}+s_d^2)+\log \alpha_d)}
\end{equation}
where $\omega^{*}$ is the posterior mean acquired from the $\omega$-step, and $s_d^2$ is the \emph{d}-th diagonal element in $S_c$. $Q_c(\alpha)$ could be regarded to obey the following Gamma distribution
\begin{equation}
\label{equ:slr_a_gamma}
Q_c(\alpha)=\prod_{d=1}^{D}{Q_c(\alpha_d)}=\prod_{d=1}^{D}{\varGamma(\alpha_d^*,\frac{1}{2})}
\end{equation}
in which $\varGamma(\alpha_d^*,\frac{1}{2})$ is the Gamma distribution with the degree of freedom being $\frac{1}{2}$ and the expectation being $\alpha_d^*$ that is
\begin{equation}
	\label{equ:slr_a_star}
	\alpha_d^*=\frac{1}{\omega_d^{*2}+s_d^2}
\end{equation}

The reformulated ARD-based sparse logistic regression with the correntropy learning framework, proposed in this paper, is named as CSLR. We then consider the optimization for $\omega$-step and $\alpha$-step, respectively. In $\omega$-step, $\log Q_c(\omega)$ is equivalent to an $L_2$-regularized MCC-based logistic regression. Although it is non-convex as a result of the integration of sigmoid function and Gaussian kernel function, it is acceptable to obtain a local optimum for $\omega$-step, because it has been proved that any local optimums of regularized $m$-estimation are sufficiently close to the global optimum \cite{loh2013regularized} while correntropy is exactly a robust formulation of the Welsch $m$-estimator \cite{liu2007correntropy}. To acquire a local optimum for $\omega$-step, one could utilize the half-quadratic (HQ) technique \cite{li2021restricted}. We provide the mathematical derivation of HQ optimization for $\omega$ in Appendix \ref{app:hq}. Considering the update for $\alpha$, one can use the following rule to accelerate the convergence
\begin{equation}
\label{equ:cslr_a_star_fast}
\alpha_d^*=\frac{1-\alpha_d^*s_d^2}{\omega_d^{*2}}
\end{equation}
which is motivated by the effective number of parameters \cite{mackay1992bayesian}.

\begin{figure*}[hbtp]
\centering
\begin{equation}
\label{equ:cslr_h}
\frac{\partial^2\log Q_c(\omega)}{\partial\omega\partial\omega^t}=\frac{1}{N\sigma^2}\sum_{n=1}^{N}{\mathbf{x}_n^t\left\{ \exp(-\frac{e_n^2}{2\sigma^2})\left[ (\frac{e_n^2}{\sigma^2}-1)y_n^2(1-y_n)^2+e_ny_n(1-y_n)(1-2y_n)\right]\right\}\mathbf{x}_n}-\varLambda
\end{equation}
\hrulefill
\end{figure*}

CSLR executes $\omega$-step and $\alpha$-step alternately, updating the model parameters and relevance parameters. During the model training, the relevance parameters of the irrelevant features can diverge to infinity, that the probability density of corresponding model parameters is distributed at zero, thus pruning irrelevant features and obtaining a sparse classifier. In practice, one could initialize each $\alpha_d$ as 1 and set an upper limit, such as $10^8$. If $\alpha_d$ exceeds the upper limit, the corresponding features will be pruned in the subsequent model training. The proposed CSLR for robust sparse classification is summarized in Algorithm \ref{cslr}.

\begin{algorithm}[h]
\caption{CSLR for Robust Sparse Classification}
\label{cslr}
\begin{algorithmic}[1]
\State \textbf{input}:

training samples $\{(\mathbf{x}_n,t_n) \}_{n=1}^{N}$;

Gaussian kernel bandwidth $\sigma$;

threshold for relevance parameter $\alpha_{\max}$;

\State \textbf{initialize}:

model parameters $\omega_d$ ($d=0,1,\cdots,D$);

relevance parameters $\alpha_d=1$ ($d=0,1,\cdots,D$);

\State \textbf{output}:

model parameters $\omega_d$ ($d=0,1,\cdots,D$)
\Repeat 
\State $\omega$-step: update $\omega_d$ according to Appendix \ref{app:hq};
\State $\alpha$-step: update $\alpha_d$ according to Equation (\ref{equ:cslr_a_star_fast});
\If{$\alpha_d\geqslant\alpha_{\max}$}
\State adjust the corresponding model parameters to zero and prune the corresponding features from the samples in the following iterations 
\EndIf
\Until the parameter change is small enough or the number of iterations exceeds a predetermined limit
\end{algorithmic}
\end{algorithm}

\section{Experiments}
\label{sec:exp}
In this section, the proposed CSLR algorithm was evaluated with a synthetic dataset, an EEG-based motor imagery dataset, and an fMRI-based visual reconstruction dataset, respectively, and was compared to the baseline, the original SLR algorithm. For the kernel-based CSLR algorithm, we utilized the five-fold cross-validation method to select the optimal kernel bandwidth from 20 values: 0.1, 0.2, 0.3, 0.4, 0.5, 0.6, 0.7, 0.8, 0.9, 1.0, 1.2, 1.4, 1.6, 1.8, 2, 4, 7, 10, 30, and 100. The thresholds for the relevance parameters were specified as $10^8$ for both CSLR and SLR algorithms. The attributes were normalized such that each feature was of mean $0$ and variance $1$ before utilizing the classification algorithms. The maximum number for iterations is set as 300 for both CSLR and SLR.

Note that, the crucial purpose of this study is to demonstrate the superior robustness of the correntropy learning method for sparse Bayesian problem, especially with the hierarchical ARD prior, which has not been investigated in the literature. Hence, the principal performance comparison shall be considered with the original ARD-based sparse model, i.e. SLR. We provide a detailed discussion for the relations between CSLR and other existing methods in Section \ref{subsec:ext_model}.

\subsection{Synthetic Dataset}
\label{subsec:toy}

\subsubsection{Dataset Description}~

First, we considered a noisy and high-dimensional synthetic dataset, with which CSLR and SLR algorithms were evaluated in respect of classification accuracy and feature selection. This example was created with a similar method as in \cite{li2021restricted}. We first randomly generated 300 i.i.d. \footnote{independent and identically distributed} training samples and 300 i.i.d. testing samples, which obeyed a 500-dimensional multivariate standard normal distribution. For the label of each sample, we generated a sparse true solution. Specifically, the true solution $\omega^*$ was a 500-dimensional vector, where only five components were relevant to the label while the other 495 components were equal to zero:
\begin{equation}
\label{equ:true_solution}
\omega^*=[ \overset{\text{500-dimensional}}{\overbrace{\omega^*_1,\omega^*_2,\omega^*_3,\omega^*_4,\omega^*_5,\underset{\text{495 components}}{\underbrace{0,0,\cdots,0}}}} ]
\end{equation}
where the non-zero components were separately subject to the univariate standard normal distribution. For each sample, the label was assigned 1 if the product between the corresponding attribute and $\omega^*$ was larger than 0, otherwise assigned 0. Thus, we would train the classifiers with 300 training samples by 500 features, and evaluate them on the other 300 testing samples.

\begin{figure}[t!]
	\centering
	\includegraphics[width=0.38\textwidth]{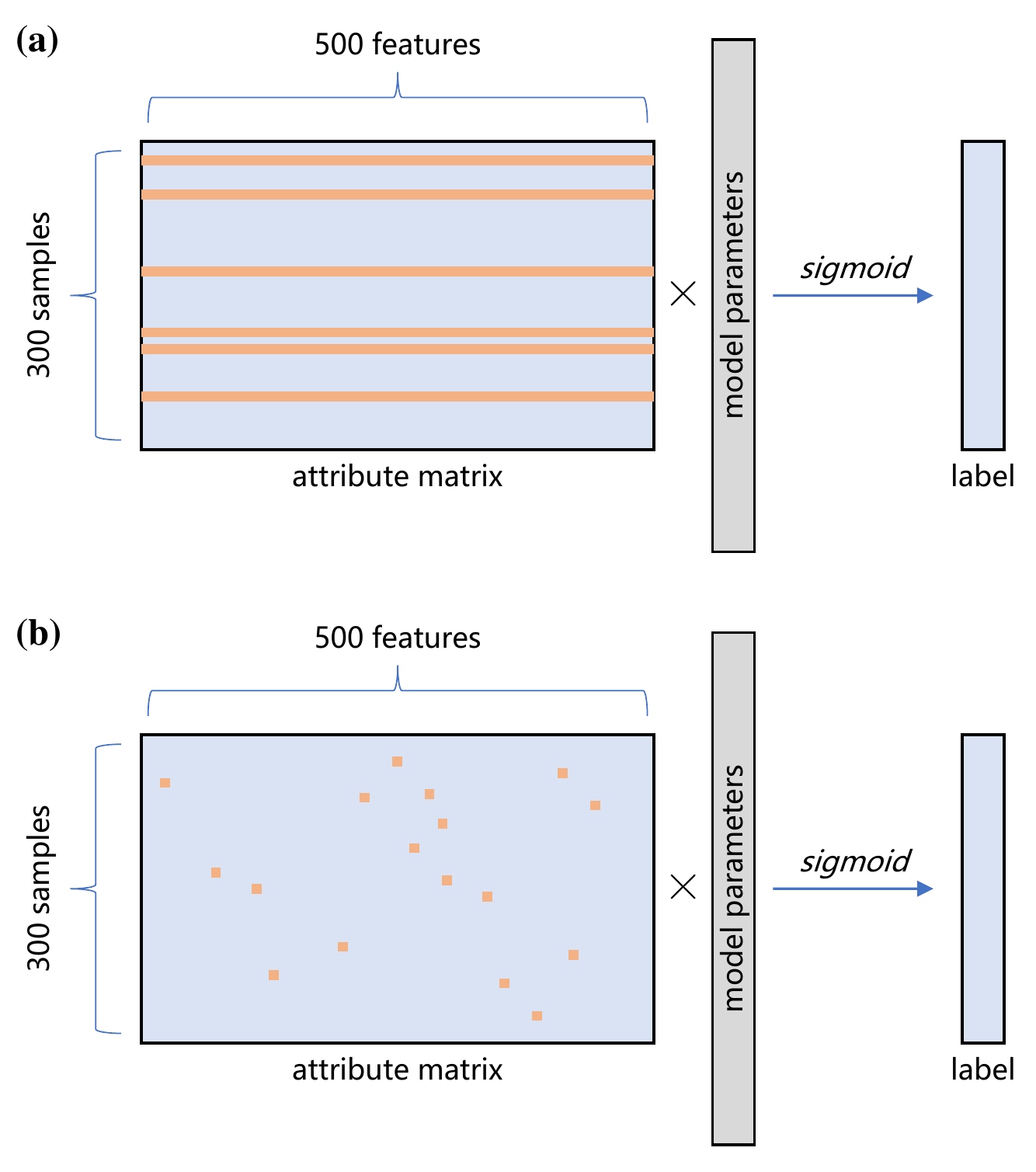}
	\caption{Two corruption models were employed for the synthetic dataset: (a) sample contamination (b) arbitrary contamination.}
	\label{fig_model}
\end{figure}

\begin{figure*}[t!]
	\centering
	\includegraphics[width=1.0\textwidth]{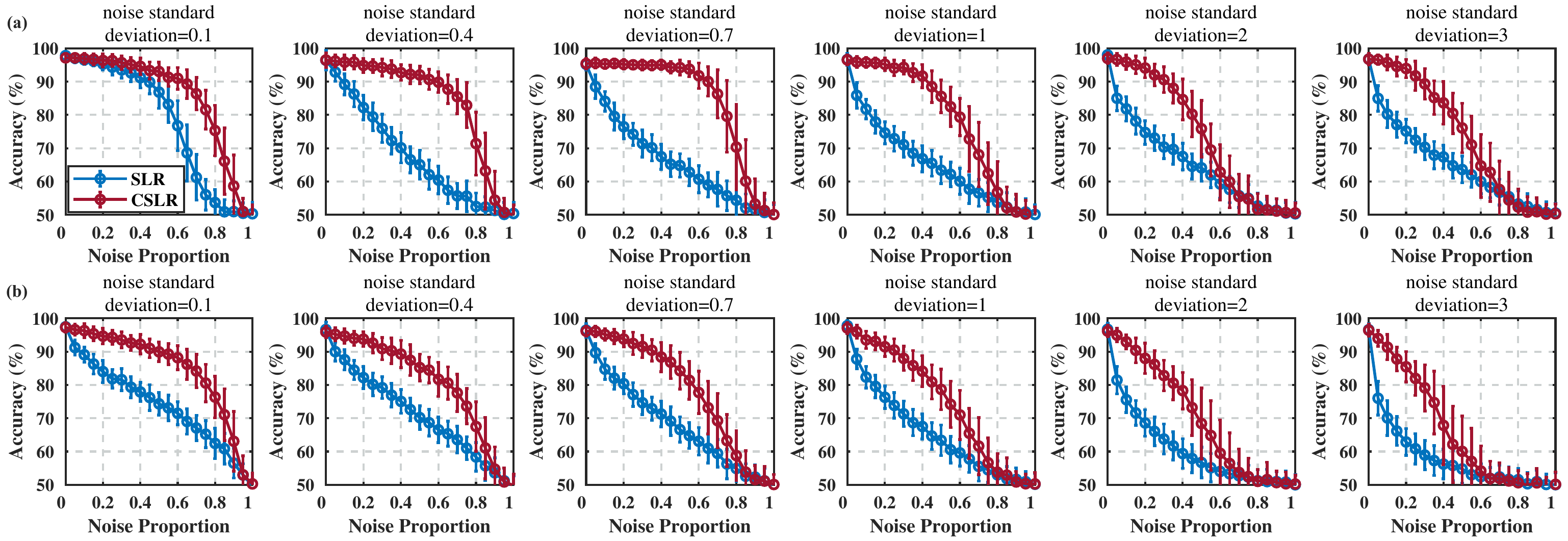}
	\caption{Classification accuracy on the noisy and high-dimensional synthetic example under two different contamination models: (a) sample contamination (b) arbitrary contamination. The results are averaged across $100$ Monte-Carlo repetitions, where the error bar indicates the corresponding standard deviation.}
	\label{fig_toy}
\end{figure*}

\begin{figure*}[t!]
	\centering
	\includegraphics[width=0.84\textwidth]{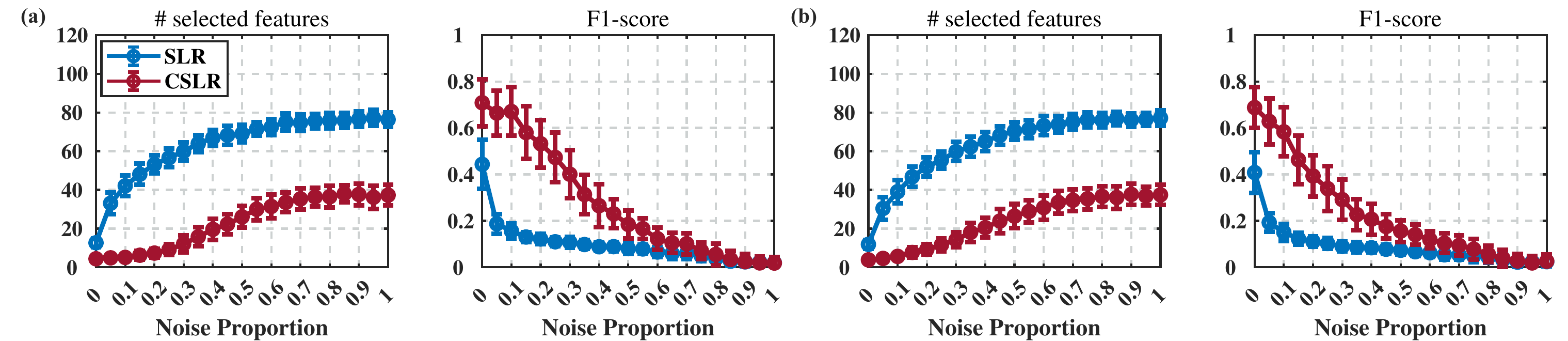}
	\caption{Number of selected features and F1-score for feature selection: (a) sample contamination (b) arbitrary contamination. The error bar denotes the standard deviation. The average results and the corresponding standard deviations are obtained from $100$ Monte-Carlo repetitions.}
	\label{fig_toy_fea}
\end{figure*}

Considering the contamination for this synthetic dataset, we utilized two different corruption models according to \cite{chen2013robust}, as shown in Fig. \ref{fig_model}. The sample contamination means undivided samples are corrupted, and the arbitrary contamination means any arbitrary elements in the attribute matrix may be corrupted. To contaminate the data, we randomly selected a certain ratio of samples or elements and replaced their attributes with zero- mean Gaussian distributed noises with the following different standard deviations: 0.1, 0.3, 0.7, 1.0, 2.0, and 3.0. To assess the robustness of different classifiers, as was suggested in \cite{zhu2004class}, one is supposed to only contaminate the training data. Hence, only the 300 training samples were corrupted in this synthetic dataset. For both sample and arbitrary contamination, the ratio of the corrupted samples/elements was increased from 0 to 1.0 with a step 0.05.

\subsubsection{Results}~

We evaluated CSLR and SLR algorithms with 100 Monte-Carlo repetitions on this synthetic dataset. First, we considered the classification accuracy, which means the proportion of the correctly classified samples in the testing dataset. The average classification accuracy for sample contamination and arbitrary contamination is illustrated in Fig. \ref{fig_toy} (a) and (b), respectively. As one could observe in Fig. \ref{fig_toy}, for both sample contamination and arbitrary contamination, the proposed CSLR outperformed the original SLR algorithm significantly when the training data suffered corruption under each noise standard deviation.  

In addition, we assessed the capability for feature selection. The feature selection can be regarded as an unbalanced binary classification, in which there were 5 relevant features and 495 irrelevant features. The sparse classifiers would select features in the model training. Hence, we could evaluate the validity of the selected features with the true `relevant'/`irrelevant' labels. We used a comprehensive performance indicator, F1-score \footnote{Denoting `relevant' as positive and `irrelevant' as negative, F1-score is the harmonic mean of the precision (TP/(TP+FP)) and the recall (TP/(TP+FN)), which is computed by 2$\times$precision$\times$recall/(precision+recall).}, to evaluate this unbalanced binary classification. Fig. \ref{fig_toy_fea} shows the number of selected features and F1-score for each algorithm in sample and arbitrary contamination, respectively. The noise standard deviation was set as 1.0. As is shown in Fig. \ref{fig_toy_fea}, CSLR always selected fewer features than SLR for both sample and arbitrary contamination. More importantly, CSLR achieved a higher F1-score for the feature selection than SLR under most noise proportions.

\subsection{EEG-Based Motor Imagery Dataset}
\label{subsec:eeg}
In addition, we evaluated the proposed CSLR algorithm on an EEG-based motor imagery decoding task with the galvanic vestibular stimulation (GVS). This dataset was proposed in our previous work \cite{shi2021galvanic}.

\subsubsection{Dataset Description}~

\begin{figure}[t!]
	\centering
	\includegraphics[width=0.42\textwidth]{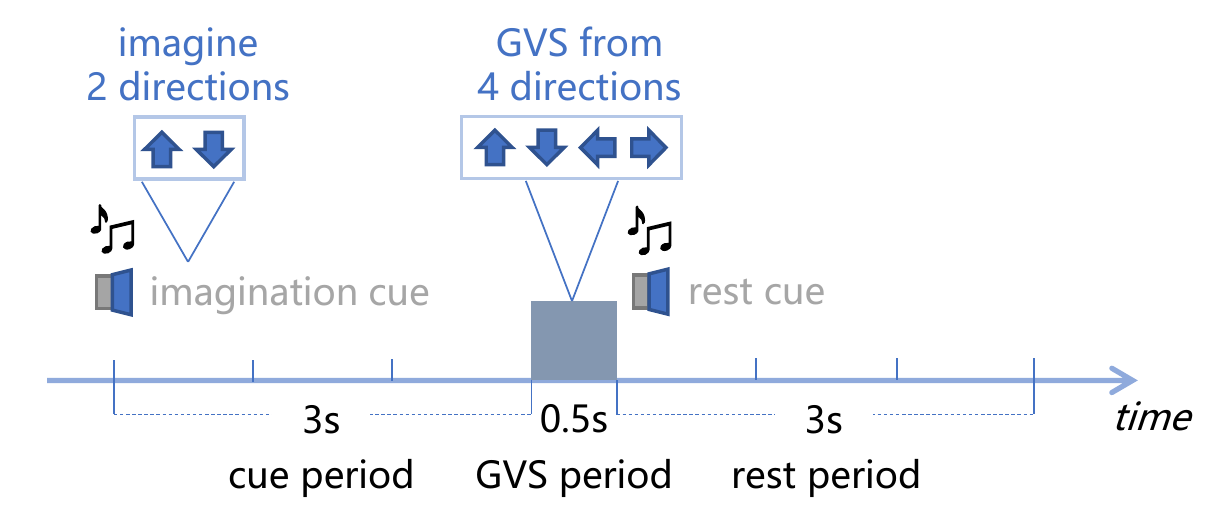}
	\caption{A schematic diagram of the EEG-based motor imagery experiment with GVS-induced sensory feedback.}
	\label{fig_eeg_diag}
\end{figure}

Ten healthy subjects were involved in this experiment. Their brain activities during the experiment were recorded by a 64-channel EEG recording system at 2048 Hz. A customized GVS instrument was used in parallel to induce the sensory feedback. In the experiment, the subjects were required to keep their eyes closed and to imagine the motions (forward/backward), with random voice-based cues. After 3-second cue period, the GVS instrument started to stimulate the subjects with four directions for 0.5 second: forward/backward/left/right. Then, the subjects rested for 3 seconds with a beep cue. A schematic diagram of one trial is illustrated in Fig. \ref{fig_eeg_diag}. Each subject participated in 6 sessions, where each session consisted of 60 trials. Thus, the data of each subject contained 360 trials. For each subject, the directions of the GVS-induced sensory feedback were identical with the imagined motion directions for 180 trials, while were inconsistent for the other 180 trials. One is supposed to predict whether the direction of motor imagery is consistent with the GVS-induced sensory feedback: match or mismatch. One can find a comprehensive description of this dataset in \cite{shi2021galvanic}.       

\subsubsection{Decoding Paradigm}~

To achieve an application for real-time BCI system, the EEG data was used for decoding analysis with a rather raw state in this dataset. The decoding paradigm is listed in what follows:

\begin{itemize}
\item[1.]
The EEG signals were down-sampled to 512 Hz.
\item[2.]
The 360 trials were reordered randomly with their labels (match/mismatch).
\item[3.]
The reordered dataset was separated, the first 80$\%$ defined as training while the remaining 20$\%$ as testing.
\item[4.]
The absolute magnitude of the EEG recording during the GVS period was used as feature for each 100 ms duration (0–0.1 s, 0.1–0.2 s, 0.2-0.3 s, 0.3–0.4 s, 0.4–0.5 s).
\item[5.]
As a result, there were 288 training samples (80$\%$) and 72 testing samples (20$\%$) with 3,264 features (64 channels $\times$ 51 samplings in 100 ms). The classifiers were trained on the training set and were evaluated on the testing set.
\item[6.]
Steps 2.–5. were implemented with 20 repetitions. Then we obtained the results for each subject.    
\end{itemize}

\subsubsection{Results}~

\begin{figure}[t!]
	\centering
	\includegraphics[width=0.48\textwidth]{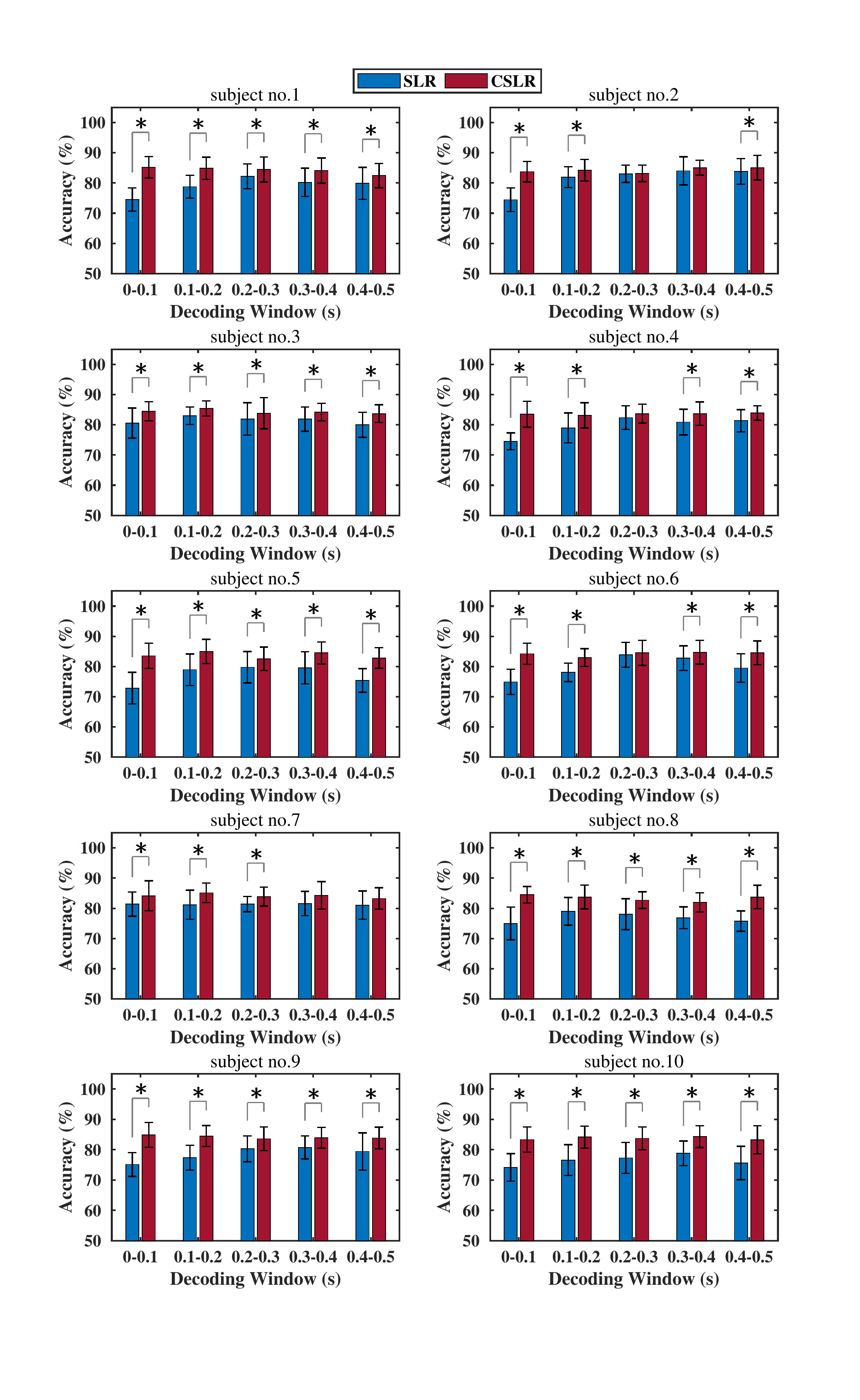}
	\caption{Classification accuracy on the EEG-based motor imagery dataset with GVS-induced sensory feedback. The results are averaged across 20 repetitions, where the whiskers denote the standard deviations. `$\ast$' indicates statistically significant difference according to a paired \emph{t}-test ($p<0.01$).}
	\label{fig_eeg}
\end{figure}

The average classification accuracy in each 100 ms decoding window is shown in Fig. \ref{fig_eeg} for each subject from 20 repetitions. The whiskers represent the corresponding standard deviations. We also examined whether there existed statistically significant difference between the accuracy obtained by CSLR and SLR, respectively, according to a paired \emph{t}-test with $p<0.01$. Among the 50 conditions in total (10 subjects $\times$ 5 decoding windows), the proposed CSLR achieved statistically higher classification accuracy in 44 conditions. Though we failed to find significant difference in the remaining 6 conditions, CSLR realized higher average classification accuracy than SLR as well. Additionally, the average classification accuracy across a total of 10 subjects in each decoding window and across all decoding windows is shown in TABLE \ref{eeg_accu}. The higher accuracy under each decoding window is marked in bold. One can observe that the proposed CSLR achieved higher average accuracy than SLR under each decoding window and across all the decoding windows.     

\begin{table*}[!tp]
\centering
\caption{Average classification accuracy across ten subjects in each decoding window and across all decoding windows with different classification algorithms.}
\label{eeg_accu}
{
\begin{tabular}{@{}ccccccc@{}} 
\toprule
\hline
\thead{Decoding \\ Window (s)} & 0-0.1 & 0.1-0.2 & 0.2-0.3 & 0.3-0.4 & 0.4-0.5 & average  \\ \hline
SLR & 75.75$\pm$5.12 &79.38$\pm$4.70 & 81.03$\pm$4.82 & 80.74$\pm$4.70 & 79.18$\pm$5.35 & 79.22$\pm$5.28 \\   
CSLR& \textbf{84.15$\pm$3.90} &\textbf{84.32$\pm$3.65} & \textbf{83.62$\pm$3.79} & \textbf{84.08$\pm$3.72} & \textbf{83.66$\pm$3.77} & \textbf{83.97$\pm$3.77} \\ \hline \bottomrule
\end{tabular}}
\end{table*}     

\begin{figure*}[t!]
	\centering
	\includegraphics[width=1.0\textwidth]{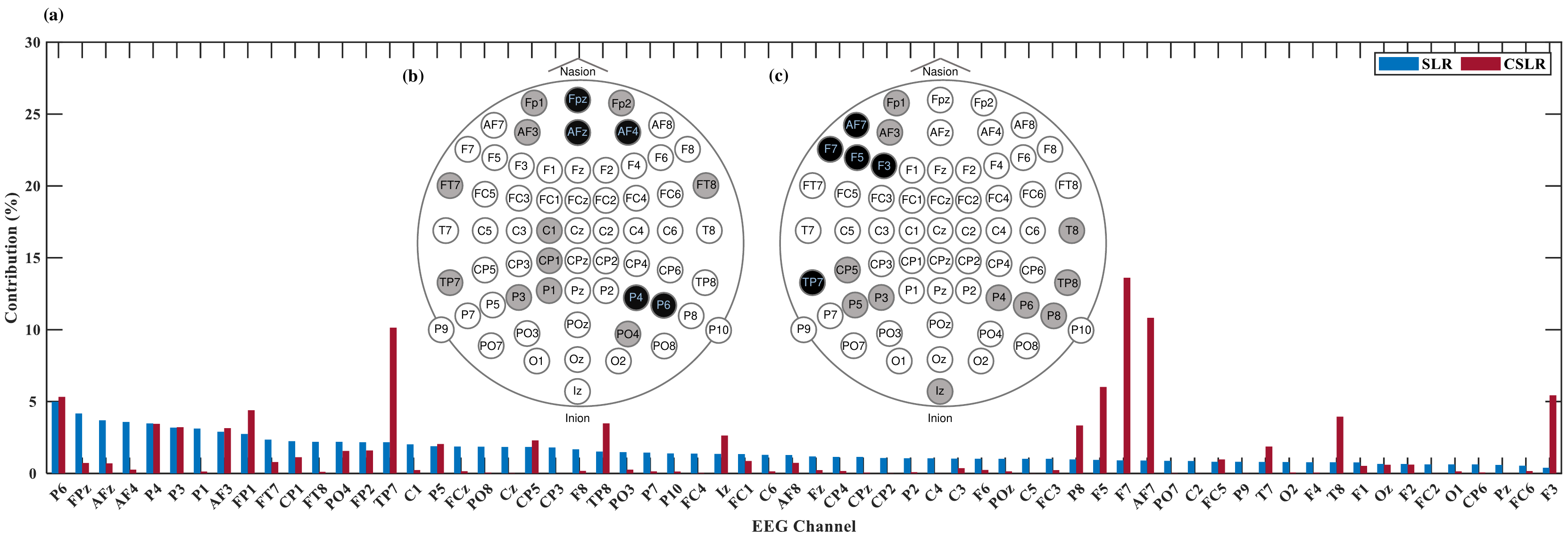}
	\caption{Spatial patterns obtained by SLR and CSLR in EEG-based motor imagery dataset with GVS-induced sensory feedback. The contribution of each EEG channel $W(\text{ch})$ is shown in (a), sorted in descending order according to the spatial contribution by SLR. The top 5 EEG channels with the maximal spatial contributions are plotted in black circles, while the other channels with the top 16 spatial contributions are plotted in gray circles for SLR in (b) and CSLR in (c), respectively.}
	\label{fig_eeg_spatial}
\end{figure*} 

\begin{table*}[!tp]
	\centering
	\caption{Average classification accuracy across ten subjects in each decoding window and across all decoding windows with the generic logistic regression algorithm and the top 5 EEG channels selected by different sparse classifiers.}
	\label{eeg_accu_chan}
	{
		\begin{tabular}{@{}ccccccc@{}} 
			\toprule
			\hline
			\thead{Decoding \\ Window (s)} & 0-0.1 & 0.1-0.2 & 0.2-0.3 & 0.3-0.4 & 0.4-0.5 & average  \\ \hline
			top 5 channels selected by SLR & 79.86$\pm$4.94 &84.04$\pm$3.72 & 83.78$\pm$3.90 & 83.64$\pm$3.75 & 83.37$\pm$4.15 & 82.94$\pm$4.37 \\   
			top 5 channels selected by CSLR& \textbf{81.10$\pm$4.07} &\textbf{84.24$\pm$3.82} & \textbf{83.84$\pm$3.89} & 83.64$\pm$3.64 & \textbf{83.69$\pm$3.91} & \textbf{83.30$\pm$4.04} \\ \hline \bottomrule
	\end{tabular}}
\end{table*}    

We further studied the spatial patterns for the classification models acquired by CSLR and SLR, respectively, calculating how much each channel contributed to the whole classification model. The element of the trained model parameter is denoted by $\omega_{\text{ch,temp}}$ which corresponds to the EEG channel `ch' and the sampling time `temp'. The spatial contribution for the channel `ch', denoted by $W(\text{ch})$, is calculated by the ratio between the summation of the absolute values of model parameters, which correspond to the current channel, and the whole classification model
\begin{equation}
\label{equ:spatial_contribution}
W(\text{ch})=\frac{\sum_{\text{temp}}{\|\omega_{\text{ch,temp}}\|}}{\sum_{\text{ch}}\sum_{\text{temp}}{\|\omega_{\text{ch,temp}}\|}}
\end{equation}
The spatial contribution for each EEG channel is illustrated in Fig. \ref{fig_eeg_spatial} (a) for SLR and CSLR, respectively, averaged across a total of ten subjects and the five decoding windows. Moreover, the top 5 EEG channels with the maximal spatial contributions are shown in black circles while the other top 16 EEG channels are presented in gray circles in Fig. \ref{fig_eeg_spatial} (b) and (c) for SLR and CSLR, respectively. To demonstrate the superior capability of feature selection for CSLR, we utilized the EEG data from the respective top 5 channels and implemented the same decoding paradigm by the MLE-based generic logistic regression model. The average classification accuracy for each decoding window across all subjects is listed in TABLE \ref{eeg_accu_chan}. One can observe that the generic logistic regression model exhibited higher average accuracy with the top 5 channels that were selected by CSLR in four decoding windows and higher average accuracy across all decoding windows.

\subsection{fMRI-Based Visual Reconstruction Dataset}
\label{subsec:fmri}
Furthermore, the proposed CSLR was assessed on an fMRI-based visual image reconstruction task, which was originally proposed in \cite{miyawaki2008visual}. This dataset is public available at Brainliner, a neuroscience data sharing platform developed by Advanced Telecommunications Research Institute International, Japan.\footnote{brainliner.jp/data/brainliner/Visual$\_$Image$\_$Reconstruction}

\begin{figure*}[t!]
	\centering
	\includegraphics[width=0.85\textwidth]{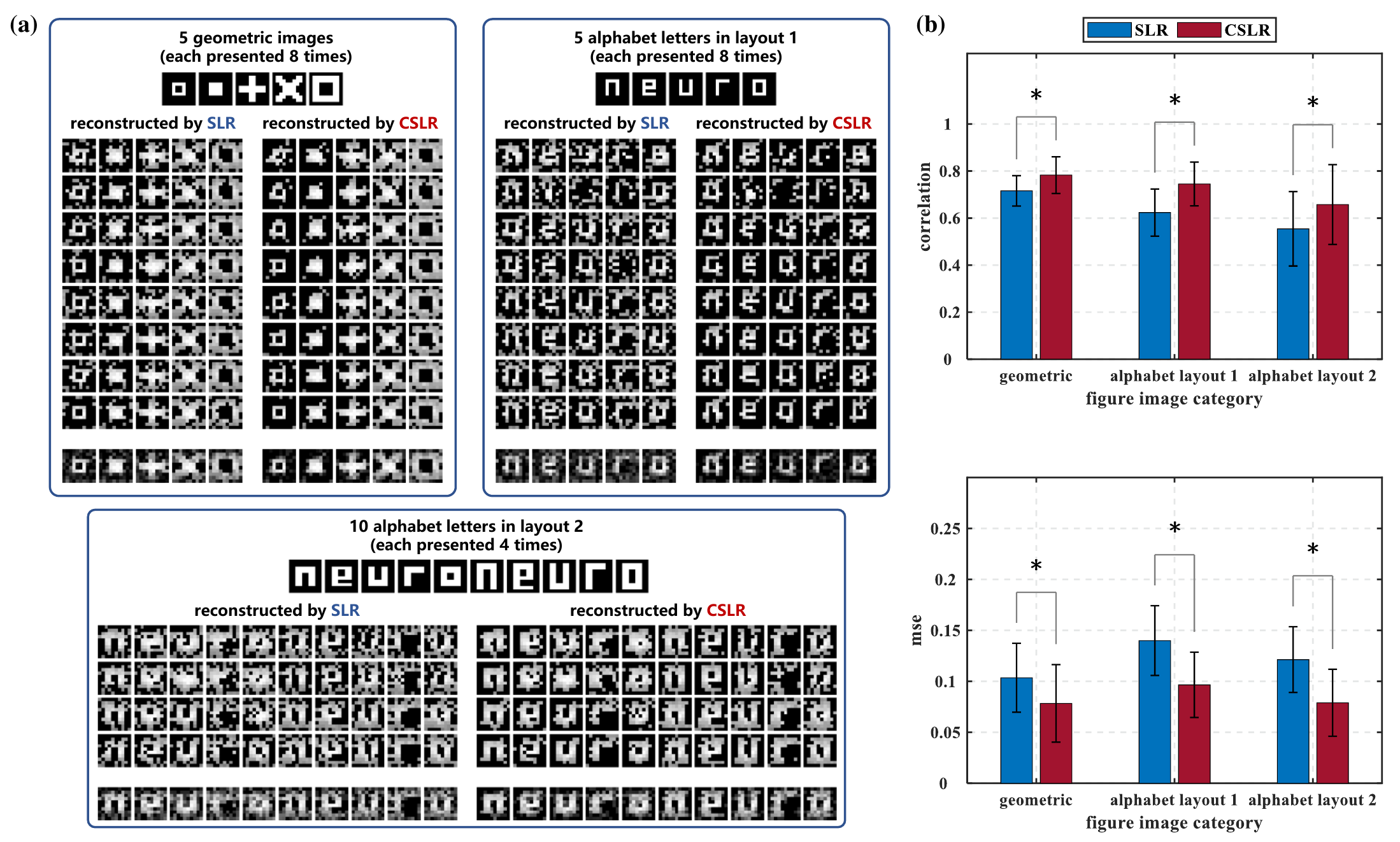}
	\caption{Diagrammatic results on the fMRI-based visual stimulus reconstruction dataset: (a) The comparison between the original and the reconstructed visual stimulus by CSLR and SLR, respectively, for three different categories in the figure image session. A total of 40 images were presented to the subject and then reconstructed by the fMRI signals for each category. The bottom rows illustrate the average reconstructed visual images for each kind of the presented figure images. (b) The spatial correlation (upper) and mean squared error (bottom) between the original and the reconstructed visual stimulus for each category, averaged across the corresponding 40 stimulus blocks. The error bars indicate the standard deviations. `$\ast$' means statistically significant difference according to a paired \emph{t}-test with $p<0.01$.}
	\label{fig_fmri}
\end{figure*}

\begin{figure}[t!]
	\centering
	\includegraphics[width=0.3\textwidth]{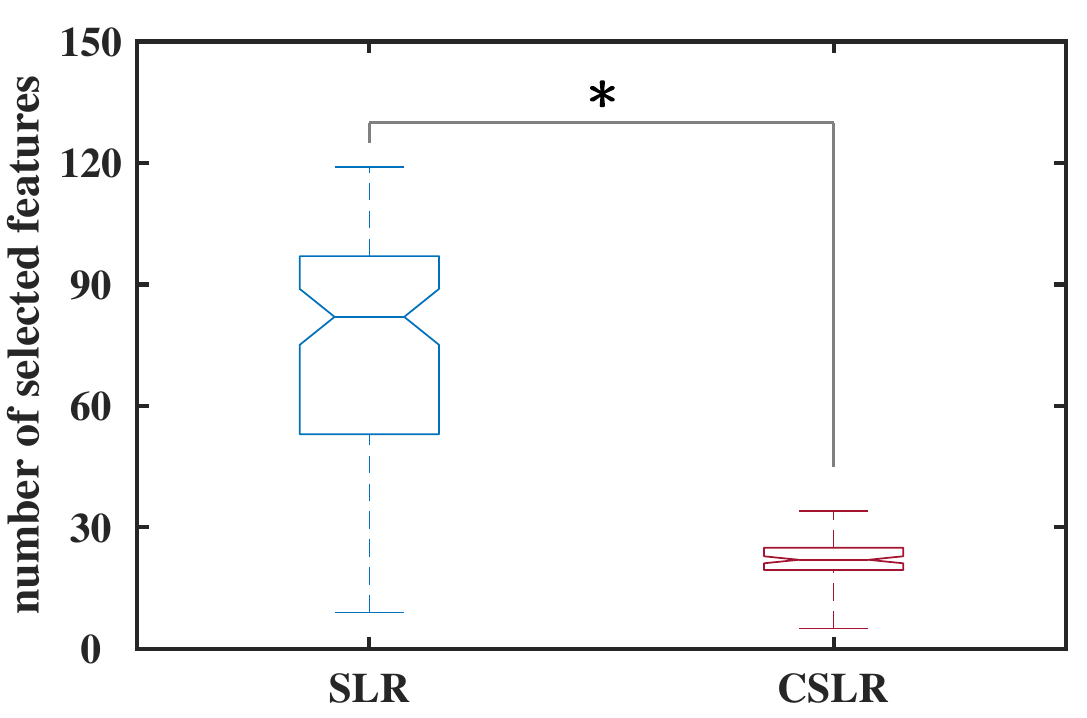}
	\caption{Number of selected features for the 100 local decoders that were ultimately trained by 440 random images. `$\ast$' means CSLR selected fewer features than SLR with statistically significant difference according to a paired \emph{t}-test with $p<0.01$.}
	\label{fig_fmri_fea}
\end{figure}

\subsubsection{Dataset Description}~

The subject was watching visual images consisting of 10 $\times$ 10 square patches. Each patch was either a homogeneous gray area or flickering at 6 Hz. Brain activities of the subject were recorded simultaneously by fMRI signals. The dataset consists of 2 sessions: one random image session and one figure image session. In the random image session, the shown images were formed in stochastic patterns. 440 different random images in total were presented to the subject. Each stimulus block lasted 6 s, followed with a 6 s rest period. In the figure image session, there were 3 types of figure images: geometric, alphabet letter layout 1, and alphabet letter layout 2. Each type had 40 blocks totally. Each stimulus block lasted 12 s, followed by 12 s rest. For geometric images, five shapes were presented 8 times. For alphabet letter layout 1, five letters were presented 8 times. For alphabet letter layout 2, ten letters were presented 4 times. The preprocessing for fMRI data was identical to the original study. In the following analysis, the V1 and V2 regions were utilized to reconstruct the images, in which 1,698 voxels in total were involved. More details of the experiment can be found in \cite{miyawaki2008visual}.

\subsubsection{Decoding Paradigm}~

The random image session was used to train reconstruction models with different classifiers while the figure image session was utilized to evaluate the reconstruction performance. As in \cite{miyawaki2008visual}, we used a linear combination of local image decoders to reconstruct the 10 $\times$ 10 images by a local image basis of size 1 $\times$ 1. Accordingly, for a 10 $\times$ 10 image, we would train 100 individual binary classifiers for each pixel, which predicted if the pixel was flickering or a gray area. Afterward, the predicted contrast values of each pixel were combined to obtain the final reconstructed images with the combination coefficients, which were acquired by 10-fold cross-validation in the random image session. A total of 440 random images were separated equally into nine training groups and one validation group. The local decoders with 100 classifiers were trained on the nine training groups. Then, we found the optimal non-negative combination coefficients by minimizing the sum of the square error between the true and the predicted validation group by the trained local decoders. The eventual combination coefficients were averaged by the cross validations. Then, we retrained the local decoders with all the 440 random images and integrated them with the combination coefficients to reconstruct the visual stimulus for the figure image session.

\subsubsection{Results}~

We assessed CSLR and SLR algorithms separately with the above-mentioned decoding method. The reconstructed images of the figure image session are illustrated in Fig. \ref{fig_fmri} (a) by each type, in comparison to the original visual stimulus. Further, we quantified the spatial correlation and mean squared error (mse) between the original and the reconstructed visual stimulus. The average result across 40 images for each figure image category is shown in Fig. \ref{fig_fmri} (b). For each category, the proposed CSLR achieved higher spatial correlation and lower mse than SLR with statistically significant difference according to a paired \emph{t}- test with $p<0.01$.

In addition, we considered the feature selection in this fMRI dataset as well. For the ultimate 100 local decoders which were trained by 440 random images to reconstruct the figure image, we counted the number of non-zero model parameters for each decoder (classifier), i.e. the number of selected features in each classification model. The result is illustrated by boxplot in Fig. \ref{fig_fmri_fea} for SLR and CSLR, respectively. One perceives that CSLR selected statistically fewer features (voxels) in fMRI decoding than SLR under a paired \emph{t}-test with $p<0.01$.

\section{Discussion}
\label{sec:disc}

\subsection{Classification Accuracy and Feature Selection}
First, we would like to give a more detailed discussion about the experimental results. There are two critical aspects to view the classification model in a high-dimensional problem, i.e. the classification accuracy and the feature selection.

The proposed CSLR was demonstrated by the experimental consequences to achieve higher classification accuracy in noisy and high-dimensional decoding tasks. In the synthetic dataset, CSLR realized nearly identical classification accuracy as SLR when there was no contamination in the data. After we added artificial contamination in the attribute matrix, especially when the noise standard deviation is larger than 0.4 for sample contamination and all the noise standard deviations for arbitrary contamination, one sees in Fig. \ref{fig_toy} that significant performance degradation happened to SLR even though the noise proportion is equal to 0.05. In contrast, the proposed CSLR realized much less performance degradation than SLR when the dataset was corrupted. Overall, in this noisy high-dimensional toy dataset, except for when the noise proportion is equal to zero or close to 1.0, the proposed CSLR almost always showed better results than SLR. Next, for the EEG data, CSLR achieved statistically higher accuracy than SLR for 44 conditions among a total of 50 scenarios (10 subjects $\times$ 5 decoding windows), while also realized higher average accuracy in the remaining 6 conditions though without significant difference (Fig. \ref{fig_eeg}). In summary, the average classification accuracy for all subjects and decoding windows was improved by 4.75$\%$ (TABLE \ref{eeg_accu}). Finally, one can observe from the fMRI-based visual reconstruction results in Fig. \ref{fig_fmri} that, the reconstructed images by CSLR are commonly more legible and closer to the original stimulus. Quantitatively, CSLR achieved higher spatial correlation and lower mse than SLR with statistically significant difference.

On the other hand, CSLR can select a more informative set of features. In Fig. \ref{fig_toy_fea}, the number of selected features by SLR is 12.66$\pm$2.25 without contamination, while is 4.40$\pm$0.96 for CSLR, which is closer to the true number of the five relevant features. Meanwhile, CSLR achieved considerably higher F1-score in feature selection (0.708$\pm$0.101 vs 0.443$\pm$0.105) than SLR. When the data was contaminated, the number of selected features and F1-score were obviously affected for SLR, while CSLR effectively suppressed the negative effects of corruption by comparison. For the EEG dataset, since it could be difficult to subjectively judge which spatial pattern is more informative as shown in Fig. \ref{fig_eeg_spatial}, we used the top 5 EEG channels selected by SLR and CSLR, respectively, and assessed them by a generic logistic regression model. Even by the same classifier, the top 5 channels of CSLR led to higher accuracy than SLR (TABLE \ref{eeg_accu_chan}), indicating that the spatial pattern of CSLR could be more informative for the decoding task. For the fMRI dataset, CSLR selected much fewer voxels than SLR, see Fig. \ref{fig_fmri_fea}. Notably we only employed the voxels in the V1 and V2 regions, which are the most informative areas for visual reconstruction, according to \cite{miyawaki2008visual}. Despite such feature selection in advance, CSLR gave better visual reconstruction results through much fewer voxels.

Note that, in the experiments of this study, we did not make any artificial contamination on the brain measurement signals, which is considerably different from the related literature such as \cite{wang2011l1,chen2018common}. We only evaluated the proposed CSLR under the original brain decoding scenarios, in which it outperformed the existing non-robust method. On the one hand, this verifies that many adverse noises could exist in the brain recording signals. On the other hand, in turn, these results further demonstrated the legitimacy to employ the correntropy learning approach in brain decoding tasks.

\subsection{Extension to Other Machine Learning Models}
\label{subsec:ext_model}
CSLR algorithm could be viewed as a preliminary proposal to integrate the correntropy learning framework with the sparse ARD technique. More generally speaking, it was proposed by combining a robust information theoretic learning criterion and sparse Bayesian learning means. To the best of our knowledge, this is the first time such a proposal has been investigated. The existing robust sparse machine learning and signal processing algorithms motivated by information theoretic learning mainly employed a sparse regularization item to a robust information theoretic learning based objective function. For example, a new robust version of the sparse representation classifier (SRC) for face recognition was proposed by using the $L_1$-regularization for the correntropy-based objective function \cite{he2010maximum,he2011regularized}. In \cite{zhou2017maximum}, a robust sparse subspace learning algorithm was proposed for unsupervised feature selection, with a regularized correntropy-based model. Moreover, a robust sparse adaptive filtering was proposed in \cite{ma2015maximum} by regularizing the MCC with the correntropy induced metric (CIM). Different from the existing studies, the proposed CSLR realizes a sparse model from the perspective of Bayesian learning with the ARD prior. A crucial advantage, as mentioned above, is one does not need to adjust the hyper-parameter manually that controls the strength of regularization term. Thus, we would like to argue that the derivation of CSLR could be intuitively generalized to the other machine learning models. For instance, the $L_1$-regularized MCC-based models can be optimized with the Bayesian learning framework, since $L_1$-regularization is equal to utilizing a Laplacian prior. More importantly, one could derive new robust sparse algorithms for other machine learning problems by employing the correntropy learning criterion and the sparse ARD prior, because the real-world data could suffer noise and numerous irrelevant features frequently.

\subsection{Extension to Multi-Class Classification}
In this study, we proposed a new robust sparse classification model by integrating the correntropy learning method with the ARD-based sparse logistic regression algorithm which focuses on the binary classification problems. The proposed CSLR can be easily generalized to the multi-class classification problems. If there are $C$ classes to be classified, each class will have an individual linear discriminant function
\begin{equation}
\label{equ:dis_function_m}
f_c(\mathbf{x},\omega^c)=\sum_{d=1}^D{\omega_d^cx_d}+\omega_0^c \quad (c=1,\cdots,C)
\end{equation}
where $\omega^c$ denotes the model parameter for \emph{c}-th class. Thus we could calculate the probability that the \emph{n}-th sample belongs to \emph{c}-th class through the \emph{softmax} function as
\begin{equation}
\begin{split}
\label{equ:p_n_m}
y_n^c\triangleq P(t_n=&c)=\frac{\exp(f_c(\mathbf{x},\omega^c))}{\sum_{k=1}^{C}\exp(f_k(\mathbf{x},\omega^k))}\\
&(c=1,\cdots,C) \\
\end{split}
\end{equation}
In multi-class problem, one usually utilizes the \emph{one-hot} coding for label expression, such as $t_n=(1,0,\cdots,0)$ if \emph{n}-th sample belongs to the first class. Similarly, one obtains the prediction by $y_n=(y_n^1,\cdots,y_n^C)$. Hence, the prediction error would be a $C$-dimensional vector by subtraction $e_n=t_n-y_n\in \mathbb{R}^C$. One can then derive the multi-class CSLR algorithm naturally with the multi-dimensional errors, which would be investigated in detail in our future works. Multi-class robust sparse classifiers could further improve the brain activity decoding performance. For example, multiscale local image decoders were proved to show better visual reconstruction results in \cite{miyawaki2008visual} than only using the decoder of 1 $\times$ 1 size, by combining 1 $\times$ 1, 1 $\times$ 2, 2 $\times$ 1, and 2 $\times$ 2 decoders. Since the decoders for the other scales necessitate multi-class classification, we merely used the 1 $\times$ 1 decoder in the evaluation. For our future works, we will assess the multi-class CSLR algorithm with the multiscale decoders.



\section{Conclusion}
\label{sec:con}
In this paper, we proposed a novel robust and sparse logistic regression algorithm, by designing a correntropy-based pseudo joint distribution with the sparse automatic relevance determination technique, which is optimized by alternate updates from Bayesian perspective. The correntropy-motivated robust sparse logistic regression algorithm, named as CSLR, was evaluated on various noisy and high-dimensional classification tasks. The experimental results demonstrated that CSLR can realize better classification accuracy and feature selection for brain activity decoding tasks. We will generalize the proposed CSLR for the multi-class situations and other machine learning algorithms in our future studies. In addition, we are working on deriving the correntropy-based likelihood function to incorporate it into the Bayesian framework with more solid theoretical guarantees.

\appendices
\section{Model Parameter Optimization}
\label{app:hq}
We aim to address the optimization issue in $\omega$-step with the half-quadratic technique. Denoting the parameter space as $\mathcal{W}$, we can rewrite $\omega$-step as
\begin{equation}
\label{equ:cslr_w}
\omega^*=\mathop{arg\max}_{\omega\in\mathcal{W}} \frac{1}{N}\sum_{n=1}^{N}{\exp(-\frac{e_n^2}{2\sigma^2})}-\frac{1}{2}\omega^t\varLambda\omega
\end{equation}
We define a convex function $\varphi(v)=-v\log(-v)+v$, in which $v<0$. The Gaussian kernel function can be expressed by
\begin{equation}
\label{equ:hq_1}
\exp(-\frac{e_n^2}{2\sigma^2})=\mathop{\sup}_{v<0}\{v\frac{e_n^2}{2\sigma^2}-\varphi(v)\}
\end{equation}
in which the supremum is achieved when and only when $v=-\exp(-e_n^2/2\sigma^2)<0$. Thus, we could rewrite (\ref{equ:cslr_w}) as
\begin{equation}
\label{equ:hq_2}
\omega^*=\mathop{arg\max}_{\omega\in\mathcal{W},v_n<0} \frac{1}{N}\sum_{n=1}^{N}{(v_n\frac{e_n^2}{2\sigma^2}-\varphi(v_n))}-\frac{1}{2}\omega^t\varLambda\omega
\end{equation}
by introducing auxiliary variables $\{v_n \}_{n=1}^{N}$. We then optimize (\ref{equ:hq_2}) by alternately updating $\omega$ and $v_n$. Given the current model parameters, one acquires $\{e_n \}_{n=1}^{N}$ with the logistic regression model. The auxiliary variables are updated by
\begin{equation}
\label{equ:hq_3}
v_n=-\exp(-\frac{e_n^2}{2\sigma^2}) \quad (n=1,\cdots,N)
\end{equation}
Then one could fix the auxiliary variables and optimize $\omega$ with
\begin{equation}
\label{equ:hq_4}
\omega^*=\mathop{arg\max}_{\omega\in\mathcal{W}} \frac{1}{N}\sum_{n=1}^{N}{v_n\frac{e_n^2}{2\sigma^2}}-\frac{1}{2}\omega^t\varLambda\omega
\end{equation}
for which one could use gradient-based optimization. Although (\ref{equ:hq_4}) remains a non-convex problem, the iterative half-quadratic approach is profitable, since we can easily guarantee the model parameters to converge to a local optimum, as proved in what follows.

Denoting the objective function in (\ref{equ:hq_2}) as $E(\omega,v_n)$ we have 
\begin{equation}
\label{equ:hq_5}
E(\omega,v_n)=\frac{1}{N}\sum_{n=1}^{N}{(v_n\frac{e_n^2}{2\sigma^2}-\varphi(v_n))}-\frac{1}{2}\omega^t\varLambda\omega
\end{equation}
By comparing the values of $E(\omega,v_n)$ for the (\emph{k}-1)-th and \emph{k}-th half-quadratic iterations, we could prove the convergence to a local optimum as
\begin{equation}
\label{equ:hq_6}
E(\omega^{k-1},v_n^{k-1})\leqslant E(\omega^{k-1},v_n^{k})\leqslant E(\omega^{k},v_n^{k})
\end{equation}
in which the first inequality is guaranteed by (\ref{equ:hq_1})(\ref{equ:hq_3}). We can also establish the second inequality provided we have a larger objective function value for (\ref{equ:hq_4}) with fixing the auxiliaries $v_n^k$. Thus, for each half-quadratic iteration, $E(\omega,v_n)$ is guaranteed to increase, until the model parameters reach a local optimum.

\ifCLASSOPTIONcaptionsoff
  \newpage
\fi
\bibliography{bibli}
\bibliographystyle{IEEEtran}
\end{document}